\documentclass{article}
\usepackage{spconf,amsmath,graphicx, caption}
\usepackage{xspace}
\usepackage{xcolor}
\usepackage{colortbl}
\usepackage{cellspace}
\usepackage{booktabs}
\usepackage{float}
\usepackage[section]{placeins}
\usepackage{enumitem}
\usepackage{tikz}
\usepackage{bbding}
\usepackage{subfigure}
\usepackage{multirow}
\usepackage{amsmath,amssymb,bm}
\usepackage{makecell}
\usepackage{url}

\usepackage{svg}
\usepackage{CJKutf8}
\usepackage{setspace}
\usepackage{multirow}
\usepackage{subfloat}
\usepackage{url}
\usepackage{color}
\usepackage{mathtools}
\usepackage{makecell}
\usepackage{paralist}
\usepackage{threeparttable}
\usepackage{arydshln}
\usepackage{soul}
\usepackage{changes}
\usepackage{tcolorbox}

\newcommand{\ignore}[1]{}

\title{Retrieval-Generation Synergy Augmented Large Language Models}

\name{
\begin{tabular}{c}
Zhangyin Feng, Xiaocheng Feng, Dezhi Zhao, Maojin Yang, Bing Qin
\end{tabular}}
\address{Harbin Institute of Technology, China}
\begin{document}

\maketitle

\begin{abstract}
Large language models augmented with task-relevant documents have demonstrated impressive performance on knowledgeintensive tasks. 
However, regarding how to obtain effective documents, the existing methods are mainly divided into two categories. One is to retrieve from an external knowledge base, and the other is to utilize large language models to generate documents. 
We propose an iterative retrieval-generation collaborative framework.
It is not only able to leverage both parametric and non-parametric knowledge, but also helps to find the correct reasoning path through retrieval-generation interactions, which is very important for tasks that require multi-step reasoning.
We conduct experiments on four question answering datasets, including single-hop QA and multi-hop QA tasks.
Empirical results show that our method significantly improves the reasoning ability of large language models and outperforms previous baselines.
\end{abstract}

\section{Introduction}
\begin{figure}[t]
    \centering
  \includegraphics[width=0.6\linewidth]{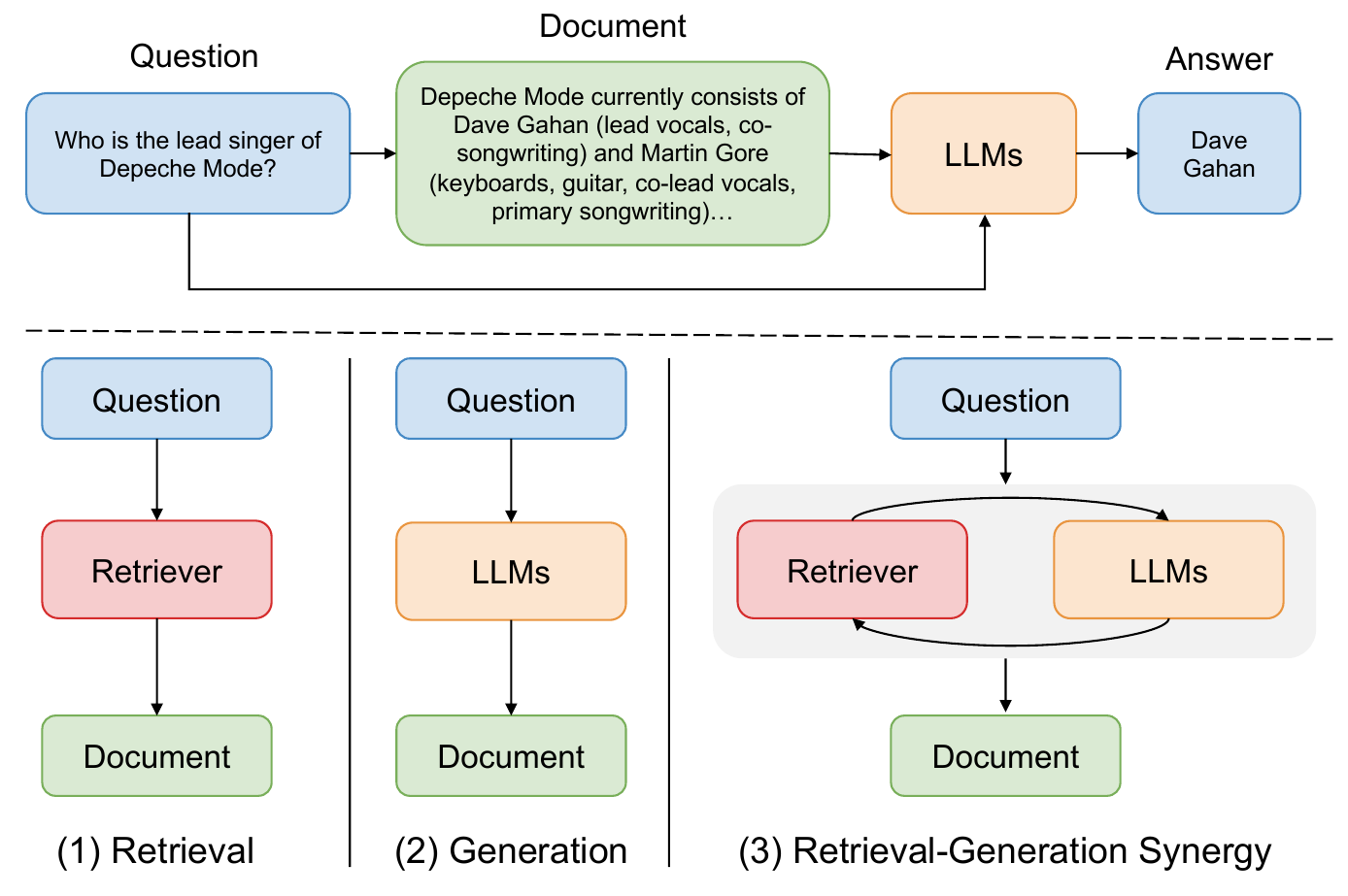}
  \caption{The top is the standard method utilizing LLMs for question answering with relevant documents. 
  The bottom shows three methods to generate relevant documents.
  }
  \vspace{-0.1cm}
  \label{fig:intro}
\end{figure}

Large Language models (LLMs) have demonstrated impressive performance on diverse language tasks through in-context learning \cite{brown2020language, hoffmann2022training, zeng2022glm, chowdhery2022palm, openai2023gpt4, touvron2023llama}.
However, they still struggle with knowledge-intensive tasks that require access to a large amount of knowledge, such as open-domain question answering \cite{lee-etal-2019-latent} and commonsense reasoning \cite{zellers-etal-2018-swag}, since the implicit knowledge preserved in the parameters may be partial and insufficient.
As shown in the top of Figure \ref{fig:intro}, one promising direction is to incorporate non-parametric knowledge to help alleviate this problem with large language models.

Recent research shows that retrieving relevant documents from an external datastore \cite{ram2023context, khattab2023demonstratesearchpredict,shi2023replug} or directly generating contextual documents from LLMs \cite{yu2023generate, sun2023recitationaugmented} both can improve LLMs' performance on knowledge-intensive tasks. 
The former, called retrieve-then-read, requires a retriever to retrieve relevant documents.
The latter, known as generate-then-read, leverages large language models to generate relevant documents before answering questions.
However, as shown in Figure \ref{fig:intro}, the above two methods are isolated and lack coordination with each other.
To fill this gap, in this paper, we explore an effective retrieval-generation collaboration framework to further improve the ability of large language models to solve knowledge-intensive tasks.

In this work, we present ITRG, an \textbf{IT}erative \textbf{R}etrieval-\textbf{G}eneration synergy framework to generate relevant documents that simultaneously exploits parametric and non-parametric knowledge.
In each iteration, ITRG consists of two important steps: generation augmented retrieval (GAR) and retrieval augmented generation (RAG).
In the GAR step, we propose a simple and effective method to expand queries by concatenating pseudo-documents generated from large language models and original questions.
And expanded queries improve the accuracy of retrieving relevant documents.
In the RAG step, we use large language models to comprehensively understand retrieved documents to generate new documents for answering questions.
We repeat these steps until we reach the maximum allowed number of iterations.
Through multiple retrieval generation collaborations, our method aids in discovering the appropriate reasoning path and providing correct answers to questions.

We evaluate the efficacy of our method on 4 question answering datasets, including Natural Questions, TriviaQA, 2WikiMultiHopQA, and HotpotQA.
Experimental results show that our method  performs better than previous baselines on all datasets.
In summary, our main contributions can be summarized as follows:
(1) We propose ITRG, an iterative retrieval-generation synergy framework using both parametric and non-parametric knowledge.
(2) We propose a simple and effective generation-augmented retrieval strategy and two retrieval-augmented generation strategies.
(3) Empirical results show that ITRG outperforms previous retrieval-augmented  methods.

\section{Iterative Retrieval-Generation Synergy}
In this section, we first introduce the overall framework, and then introduce the retrieval-generation collaboration framework in detail, including generation augmented retrieval and retrieval augmented generation.

\subsection{Overview}
\begin{figure}[t]
    \centering
  \includegraphics[width=1\linewidth]{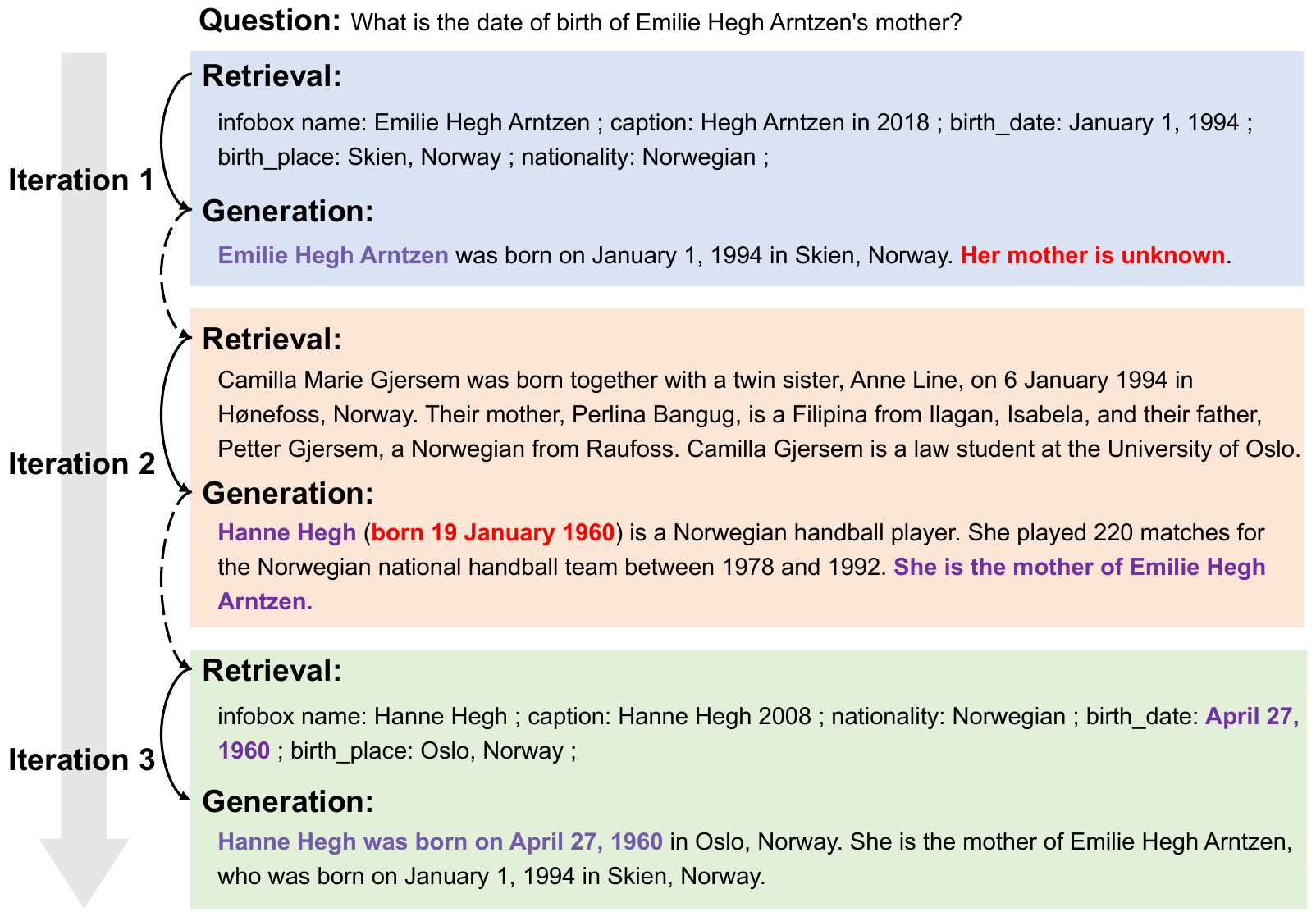}
  \caption{Iterative retrieval-generation synergy framework contains two steps in each iteration: (1) generation augmented retrieval (GAR): utilize the output of the previous iteration to expand the query to help retrieve more relevant documents; (2) retrieval augmented generation (RAG): utilize retrieved documents to generate new documents to answer questions.
    We only show three iterations in this figure for brevity. 
  Solid arrows indicate RAG within an iteration, and dashed arrows indicate GAR between iterations.
  Purple represents correct and useful information, and red represents wrong or invalid information.}
  \label{fig:method}
\end{figure}

We show the framework of ITRG in Figure \ref{fig:method}.
Given a user question $q$ and a document corpus $\mathcal{D}=\{d_i\}^{|\mathcal{D}|}_{i=1}$ (i.e, 
$d_i$ is a Wikipedia paragraph.), ITRG repeats generation augmented retrieval (GAR) and retrieval augmented generation (RAG) for $T$ iterations.
In the GAR process of iteration $t$, we concatenate the output $y_{t-1}$ of the last iteration and question $q$ to form a new query, and then use a dense retriever to retrieve top-$k$ paragraphs.
In the first iteration, we only use the question as the query.
In the RAG process of iteration $t$, based on the question $q$ and the retrieved top-$k$ paragraphs, we exploit large language models to generate new paragraphs to answer questions.
Specifically, we propose two methods to generate new paragraphs, which will be introduced in detail in \S \ref{RAG}.

\subsection{Generation Augmented Retrieval}
\label{GAR}
Knowledge-intensive tasks (e.g., open-domain question answering) often require access to additional documents. 
A common approach is to directly employ the question as the query, and then equip a sparse or dense retriever to retrieve relevant documents.
In practice, we find that in some cases using the question directly as the query fails to retrieve relevant documents because there may exist semantic gaps between them.
To alleviate this problem, we propose a simple query expansion method.
At the first iteration ($t=1$), we use the original question $q$ as the query. At iteration $t$ ($t>1$), we concatenate the original question $q$ and the document  generated $y_{t-1}$ in the last iteration as the new query $q_{t} = [q;y_{t-1}]$.
Then, we utilize a pre-trained dense retriever to retrieve top-$k$ documents, which are denoted as $R_{t}=\{d\}.$

Given an input question $q$, the retriever aims to retrieve a small set of documents from a corpus $\mathcal{D}=\{d_i\}^{|\mathcal{D}|}_{i=1}$ that are relevant to $q$.
Following prior work \cite{izacard2020leveraging}, we use a dense retriever based on the dual encoder architecture, where an encoder is used to encode both the input context $q$ and the document $d$. 
Specifically, the encoder maps each document $d \in \mathcal{D}$ to an embedding $\mathbf{E}(d)$ by taking the mean pooling of the last hidden representation over the tokens in $d$. 
At query time, the same encoder is applied to the input context $q$ to obtain a query embedding $\mathbf{E}(q)$. The similarity between the query embedding and the document embedding is computed by their cosine similarity: $s(d, q) = \cos (\mathbf{E}(d), \mathbf{E}(q))$.
The top-$k$ documents that have the highest similarity scores are retrieved.

\subsection{Retrieval Augmented Generation}
\label{RAG}
Following previous work \cite{sun2023recitationaugmented}, for a given question $q$, we could directly prompt large language models to generate related documents without retrieving them from an external corpus.
However, we find that if only the parametric knowledge learned by the large model in the pre-training stage is used, the generated documents may be incomplete.
Retrieval augmented generation (RAG) aims to comprehensively understand the retrieved non-parametric knowledge and the parametric knowledge inside large language models to generate more accurate factual knowledge.
Specifically, we propose two strategies, which will be described in detail below.

\subsubsection{Refine}
An intuitive idea is to refine the previously generated document $y_{t-1}$ based on the original question $q$ and the retrieved top-$k$ documents at the current iteration step $R_t$ to obtain a new document $y_t$. 
We call this method refine.
Considering that the document retrieved in the last iteration $R_{t-1}$ has been used to generate the last document $y_{t-1}$, 
we refine the previous output $y_{t-1}$ with updated documents $R_{update}$.
\begin{gather}
    R_{update} = R_t - R_{t-1},  \\
    y_{t} = \mathcal{M}\left(\operatorname{prompt}\left(y_{t-1}, q, R_{update}\right)\right),
\end{gather}
where $R_{update}$ means that these documents are only retrieved in the current iteration, not in the last iteration, $ \mathcal{M}$ denotes a well pre-trained large language model.
If $R_{update}$ is an empty set, we do not regenerate a new document and set $y_t=y_{t-1}$.

\subsubsection{Refresh}
In order to avoid the negative effect of errors or hallucinations in the previously generated document $y_{t-1}$, we do not use $y_{t-1}$, which is used in refine.
We refresh the memory and let the large language models directly generate the document $y_t$ based on the retrieved document $R_t$ and the original question $q$. This method is named refresh.
\begin{eqnarray}
y_{t} & = & \mathcal{M}\left(\operatorname{prompt}\left(q, R_t\right)\right)
\end{eqnarray}

Both refine and refresh are implemented through prompts. We give the prompt corresponding to refresh.
\begin{tcolorbox}[colback=black!1!white,colframe=black!57!white,title=Prompt for refresh  with all documents]
In the following task, you should write a document that contains the answer to the question.\\
\\
Passage: \{$R_{t}$\} \\
Question: \{$q$\}\\
Document: \{$y_t$\}
\end{tcolorbox}

\section{Experimental Setup}
\subsection{Datasets}
We evaluate the effectiveness of ITRG on four open domain question answering datasets, including Natural Questions  (NQ) \cite{kwiatkowski-etal-2019-natural}, TriviaQA \cite{joshi-etal-2017-triviaqa}, 2WikiMultiHopQA \cite{ho-etal-2020-constructing}  and HotpotQA \cite{yang-etal-2018-hotpotqa}. 
Following previous works \cite{trivedi2022interleaving, jiang2023active}, we randomly sub-sample 500 examples from each dataset due to the cost of running experiments.
We evaluate our method in 0-shot, 1-shot and 5-shot settings.
The few-shot demonstrations are randomly sampled from the data that is not involved in the evaluation process.

\subsection{Baselines}
\textbf{GPT-3.5} \cite{ouyang2022training}
We use text-davinci-002 and text-davinci-003 as our baselines.
Text-davinci-002 is an InstructGPT model  while Text-davinci-003 is trained with reinforcement learning with reward models trained from comparisons by humans.
\textbf{Vanilla LM} The vanilla LM baselines prompt an LLM to directly generate an answer following the few-shot in-context learning paradigm \cite{brown2020language}. 
\textbf{CoT} 
We follow \cite{wei2022chain} to generate both the chain-of-thought (CoT) reasoning process and the final answer. We only evaluate this method on multi-hop reasoning datasets in 5-shot setting\footnote{We also conduct evaluation in 1-shot setting, but the final answer could not be generated according to the corresponding instructions}.
\textbf{Retrieve-then-Read}
The retrieve-then-read baseline consists of a well-pre-trained dense retriever and a large language model.
The retriever retrieves relevant documents for the question, and then the LLM conditions on both the question and retrieved documents to generate the answer.
\textbf{Generate-then-Read}
Generate-then-read baseline first uses few-shot prompts to generate a question-related document, and then concatenates it with the question to regenerate the answer.

\subsection{Details}
LLaMA \cite{touvron2023llama} is an open source well trained large language model.
Considering the performance and computational cost of the model, we use LLaMA 33B as the backend LLM.
We use greedy decoding for both document generation and answer generation, and set up to generate 200 tokens and 15 tokens respectively.
We retrieve the top-5 paragraphs for each query and set the maximum number of iterations $T$ to 5.
We directly use the pre-trained dense retriever \cite{izacard2022few} and used the December 2018 Wikipedia dump as the retrieval corpus for all datasets.
Generated answers are evaluated with the standard exact match metric (EM score): a generated answer is considered correct if it matches any answer of the list of answers after normalization.
For this normalization step, we lowercase generated answers and remove articles, punctuation and duplicate whitespaces.

\section{Results}
\subsection{Main Results}
\begin{table*}[ht]
  \centering
  \small
    \caption{
   Exact match performance on single-hop question answering. All ITRG results are from the last iteration ($T=5$). }
   \begin{tabular}{llcccccc}
    \toprule
          & \multirow{2}{*}{Method} & \multicolumn{3}{c}{Natural Questions}  & \multicolumn{3}{c}{TriviaQA} \\
          \cmidrule(lr){3-5} \cmidrule(lr){6-8}  
          & & 0-shot & 1-shot & 5-shot & 0-shot & 1-shot & 5-shot  \\
    \midrule
    \multirow{2}{*}{GPT 3.5}
    & Text-davinci-002 & 12.0 & 24.6 & 33.0 & 46.0 & 74.2 & 76.0\\
    & Text-davinci-003 & 29.4	& 33.0	& 33.8 &	75.8 &	78.6	&77.8 \\
    \midrule
    \multirow{6}{*}{LLaMA 33B}
    &Vanilla LM     &  27.0 &	29.4&	32.4 & 74.8&	70.8&	75.8   \\
    &Retrieve-then-Read  &  27.8	 & 30.6	 & 29.8  & 74.6 &	76.0 &	76.0\\
    &Generate-then-Read     &28.0	 &31.4	 &31.0 	&73.6	&77.2&	77.6 \\
    & ITRG (refine) & 34.4 & 34.6 & 34.8 & \textbf{79.0} & \textbf{79.4} & \textbf{80.6} \\
    & ITRG (refresh)  & \textbf{37.6}	 & \textbf{38.4}	 & \textbf{38.0} & 77.0  &	78.6 &	79.4  \\
    \bottomrule
  \end{tabular}

  \label{tab:single}
 
\end{table*}
\begin{table*}[ht]
  \centering
  \small
    \caption{
    Exact match performance on multi-hop question answering. All ITRG results are from the last iteration ($T=5$). }
   \begin{tabular}{llcccccc}
    \toprule
          & \multirow{2}{*}{Method} & \multicolumn{3}{c}{2WikiMultiHopQA} & \multicolumn{3}{c}{HotpotQA} \\
          \cmidrule(lr){3-5} \cmidrule(lr){6-8} 
          & & 0-shot & 1-shot & 5-shot & 0-shot & 1-shot & 5-shot  \\
    \midrule
    \multirow{2}{*}{GPT 3.5}
    & Text-davinci-002 & 16.4&	27.6&	30.8	&12.2	&20.2	&22.2\\
    & Text-davinci-003  &  27.2 &	27.0 & 29.8 & 25.0 &	25.8 & 26.6\\
    \midrule
    \multirow{8}{*}{LLaMA 33B}
    &Vanilla LM    &  24.4&	27.6	&31.8&	22.6	&25.0	&27.0   \\
    & COT & - &-& 32.2 & - & - & 28.6 \\
    &Retrieve-then-Read & 27.4	& 29.2 &	32.0	&28.4	&29.8&	30.4     \\
    &Generate-then-Read & 30.0	& 30.4 &	31.6&	25.0&	27.0&	27.0   \\
    & ITRG (refine) &  \textbf{33.0} &	33.6 &	37.0 & 	28.8	& 29.6	&30.6  \\
       & ITRG (refresh)  &   32.2 & \textbf{36.2} & \textbf{38.6} & \textbf{31.0} &	 \textbf{32.6}	& \textbf{33.4} \\
    \bottomrule
  \end{tabular}

  \label{tab:multi}

\end{table*}

Table \ref{tab:single} reports the results on the single-hop question answering datasets. 
In the 1-shot and 5-shot settings, the performance of LLaMA-33B based Vanilla LM is very close to that of text-davinci-003. 
This shows LLaMA-33B is a strong language model, and it is reasonable to choose LLaMA-33B as our backend LLM.
Retrieve-then-read and generate-then-read all exceed vanilla LM, verifying that adding relevant external knowledge can improve the reasoning ability of large language models.
In addition, we observe that our iterative retrieval-generation collaborative method ITRG achieves state-of-the-art performance on both datasets.
Specifically, ITRG (refresh) performs better on the NQ dataset, and ITRG (refine) performs better on the TriviaQA dataset.

\begin{table}[ht]
 \small
  \centering
    \caption{Exact match performance of ITRG (refresh) at different iterations in 5-shot setting.}
   \begin{tabular}{lccccc}
    \toprule
         Iteration & 1 & 2 & 3 & 4 &5  \\
    \midrule
    Natural Questions & 34.0 & 35.2 & 37.0 & 37.2 & 38.0 \\
    TriviaQA & 79.8 & 79.2 & 79.8 & 79.8 & 79.4\\
    2WikiMultiHopQA & 34.8 & 37.4 & 37.2 & 38.6 & 38.6\\
    HotpotQA & 32.6 & 32.8 & 34.0 & 33.4 & 33.4\\
    \bottomrule
  \end{tabular}
   \vspace{-0.1cm}
  \label{tab:iter_em}
\end{table}

Table \ref{tab:multi} presents the results on the multi-hop question answering datasets.
We observe that LLaMA-33B is still comparable to text-davinci-003 on the multi-hop question answering datasets.
In addition, CoT can answer questions more accurately than vanilla LM  by generating reasoning process.
Compared with different baseline models, ITRG significantly improves the exact match scores.
Specifically, on the 2WikiMultiHopQA dataset, the exact match  score of ITRG (refresh) in the zero-shot setting is 32.2, which exceeds the performance of vanilla LM in the 5-shot setting with a score of 31.8.
In the 5-shot setting, ITRG (refresh) achieves 38.6 EM score and  improves by 6.8 points in absolute gains.
Compared to vanilla LM, ITRG (refresh) can improve the EM score by 9.4, 7.6, and 6.4 points respectively in 0-shot, 1-shot, and 5-shot settings on the Hotpotqa dataset.

\subsection{Performance at Different Iterations}
In this section, we analyze the performance of our model and the quality of the generated documents during the iteration process.
Specifically, we present the results of ITRG (refresh) at different iterations in 5-shot setting in Table \ref{tab:iter_em}.
We measure the answer recall of generated documents at different iteration steps and present results in Table \ref{tab:answer_recall}.
Table \ref{tab:iter_em} shows that the performance of the model gradually improves with iteration.
\begin{table}[ht]
 \small
  \centering
    \caption{
  Answer recall of generated documents at different iterations with ITRG (refresh).}
   \begin{tabular}{lccccc}
    \toprule
    Iteration & 1 & 2 & 3 & 4 &5  \\
    \midrule
    Natural Questions & 44.0 & 46.4 & 48.4 & 48.8 & 48.0 \\
    TriviaQA & 18.8 & 19.0  & 20.2 &19.2 &19.2\\
    2WikiMultiHopQA & 34.2 &36.6&35.0 &40.0 & 37.0\\
    HotpotQA & 34.2 & 34.8 & 35.6 & 33.8 & 33.6\\
    \bottomrule
  \end{tabular}
    \vspace{-0.1cm}
  \label{tab:answer_recall}
\end{table}
And Table \ref{tab:answer_recall} shows that the quality of the generated documents also gradually improves with iteration.
These results verify that our iterative retrieval-generation collaborative framework is effective and can further enhance the reasoning capabilities of large language models.

\section{Conclusion}
In this paper, we present ITRG, which is an iterative retrieval-generation synergy framework, containing two important steps: generation-augmented retrieval and retrieval-augmented generation.
They form a closed loop, and can improve each other via multiple iterations.
We propose a simple and effective generation-augmented retrieval strategy and two retrieval-augmented generation strategies.
Empirical results show our approach significantly exceeds several strong baselines, including GPT 3.5, on four open domain question answering datasets,
which indicates that our method can significantly improve the reasoning ability of large language models.

\bibliographystyle{IEEEtran}
\ninept
\bibliography{ref}

\end{document}